# A Comprehensive Dataset for Underground Miner Detection in Diverse Scenario

Cyrus Addy[1][0009-0001-4483-6526], Ajay Kumar Gurumadaiah[2][0000-0003-1182-7241], Yixiang Gao[3][0000-0001-8275-9069] and Kwame Awuah-Offei[4][0000-0001-5363-7492]

[1,2,3,4] Department of Mining and Explosive Engineering
Missouri University of Science and Technology
Rolla, USA Princeton University, Princeton NJ 08544, USA

`ca8mc@mst.edu, agzgx@mst.edu, ygao@missouri.edu, and kwamea@mst.edu`

**Abstract.** Underground mining operations face significant safety challenges that make emergency response capabilities crucial. While robots have shown promise in assisting with search and rescue operations, their effectiveness depends on reliable miner detection capabilities. Deep learning algorithms offer potential solutions for automated miner detection, but require comprehensive training datasets, which are currently lacking for underground mining environments. This paper presents a novel thermal imaging dataset specifically designed to enable the development and validation of miner detection systems for potential emergency applications. We systematically captured thermal imagery of various mining activities and scenarios to create a robust foundation for detection algorithms. To establish baseline performance metrics, we evaluated several state-of-the-art object detection algorithms including YOLOv8, YOLOv10, YOLO11, and RT-DETR on our dataset. While not exhaustive of all possible emergency situations, this dataset serves as a crucial first step toward developing reliable thermal-based miner detection systems that could eventually be deployed in real emergency scenarios. This work demonstrates the feasibility of using thermal imaging for miner detection and establishes a foundation for future research in this critical safety application.

## 1   Introduction

Human detection is a crucial and complex task in applications such as surveillance monitoring and emergency response. This detection increases the ability to protect life and property, strengthening security in many different scenarios [1] . The procedure covers linked tasks including sensor choice, dataset collecting, annotation, machine learning method, and evaluation metric selection [2]. The effectiveness of human detection algorithms is greatly affected by the availability of high-quality, large-scale datasets [3].

Detecting humans in underground mines is even more challenging because of environmental factors such as low lighting [4]. Current datasets for human detection underground are either limited in volume or lacking in diversity concerning miner postures and activities because of the challenges associated with data collection underground [5], [6], [7]. The scarcity hinders the development of effective miner detection models and transfer learning applications, which are crucial for enhancing emergency response and reducing human casualties in underground mines.

Thermal imaging is more suitable for underground environments than conventional RGB imaging because thermal images offer reduced noise, consistent image quality irrespective of lighting conditions, and clear separation between individuals and their surroundings due to heat signatures [8]. Moreover, thermal imaging protects volunteer privacy by making it difficult to discern personally identifying traits [9]. Thermal imaging is a preferred choice for subterranean uses since past research shows it to be effective for human detection [10].

There is, however, a significant lack of publicly accessible thermal image datasets for the detection of underground miners. The limited availability of such datasets complicates the training and assessment of deep learning models for miner identification, which is essential for improving safety and emergency response in underground mines. The absence of an adequate dataset poses significant obstacles in the development of effective detection models and the execution of transfer learning [3], [11].

Therefore, this study has two goals. First, we introduce a new miner detection dataset, Thermal Underground Human Detection (Thermal UHD) dataset, comprising thermal images obtained from an underground mining setting. This dataset encompasses various mining operations, including active work, rest, and emergency situations (e.g., fluctuations in fire, smoke, and heat). Second, by training object detection models such as YOLOv8, YOLOv10, YOLOv11, and RT-DETR, we demonstrate the dataset's suitability for transfer learning. Our objective is to establish a foundation for future breakthroughs in this domain and enhance understanding of underground miner detection.

This paper is divided into six (6) sections. Section one (1) introduces the paper, section two (2) presents an overview of related thermal datasets, section three (3) presents the hardware employed and data collection setup used to generate the dataset, section four (4) presents the object detection benchmarks using transfer learning, section five (5) presents the transfer learning using our dataset, and section six (6) presents the conclusions to this paper and recommendations for future research.

## 2    Related Thermal Datasets

Several thermal image datasets have been developed for various applications, ranging from pedestrian detection in urban environments to wildlife monitoring [1], [3], [8], [12], [13], [14], [15], [16], [17]. Developing a thermal image dataset requires a correct understanding of the domain for which the dataset is to be developed. Factors such as visibility, weather conditions, logistics, and volunteers should be considered when generating a dataset [15].

The existing thermal datasets are centered around terrestrial applications such as surveillance, tracking, and detection. However, these datasets are not tailored to the unique conditions of underground mines, such as low visibility. The Thermal UHD dataset fills this gap by providing thermal images specifically captured in a subterranean mining context, thereby offering a more relevant benchmark for developing miner detection algorithms.

## 3  Hardware and Data Collection Setup

### 3.1  Hardware

The data collection for the Thermal UHD dataset was conducted using Spot CAM+IR with a ROS enabled laptop designed to operate effectively in harsh underground conditions. The camera was equipped with high-resolution thermal sensors capable of detecting miners varying temperature. Data was collected in an active mining site, capturing various scenarios such as miners at different distances, angles, and in varying postures.

The experimental data collection setup also accounted for environmental factors like ambient temperature and humidity, ensuring the dataset's applicability to real-world mining conditions. The hardware specifications of equipment employed for this experiment are captured in Table 1.

Table 1. Hardware specifications.

| Equipment | Specification |
|---|---|
| Spot CAM+IR | a. Pan-tilt-zoom (PTZ) camera with 30× zoom |
|  | b. Temperature ranges from $-40^oC\ to\ +550^oC$ |
|  | c. Display 720×575 pixels |
| Laptop | Dell latitude 13th Gen Intel ® Core™ i7 |
|  | Ubuntu 20.04 |
|  | ROS Neotic |

### 3.2  Data collection setup

Due to the nature of the experiments which involve human subjects, the authors sought and obtained approval from the University of Missouri Institutional Review Board. All experiments were conducted at the Missouri University of Science & Technology Experimental Mine, located at 12350 Private Drive 7002, Rolla, Missouri. The Experimental Mine has two underground and two surface mines with a building complex on a total land size of about 20 acres. Fig. 1 shows sample of images collected during the experiment. Fig. 2 shows the hardware setup used for the thermal image dataset collection.

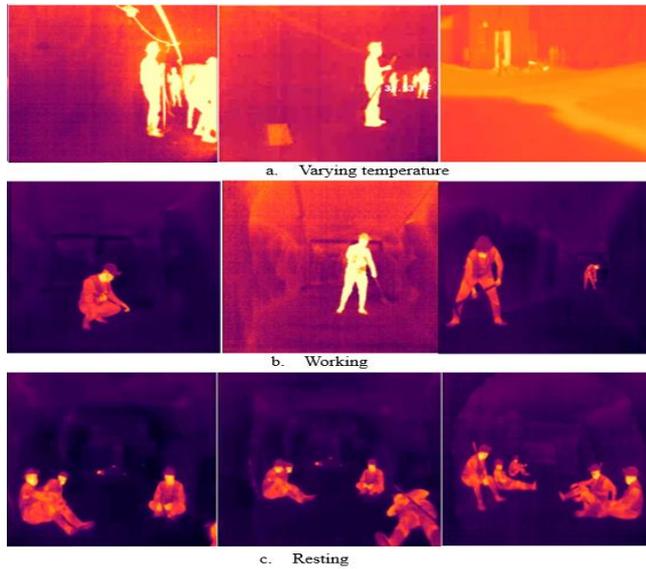

**Fig. 1.** Sample images generated from the experiments corresponding to the three phases during our data collection.

### 3.3    Data diversity and labels

Emergencies in an underground mine could be in various forms. To ensure that most of them can be captured, we employed three phases of data collection to achieve a diversified dataset, which included normal and emergency scenes. Phase 1 as depicted in Fig. 1-a required miners to be working while a varying heat source is introduced into the working environment. Also, as part of phase 1, some images were captured using fog machines to represent smoke in the underground mine. Phase 2 as shown in Fig. 1-b involved miners in working conditions under normal temperature while holding various tools. Phase 3 as shown in Fig. 1-c required miners to be in resting condition. Overall, the dataset covers five main posture/object classes, which are lying, bending, sitting, squatting, and standing for the various scenarios considered.

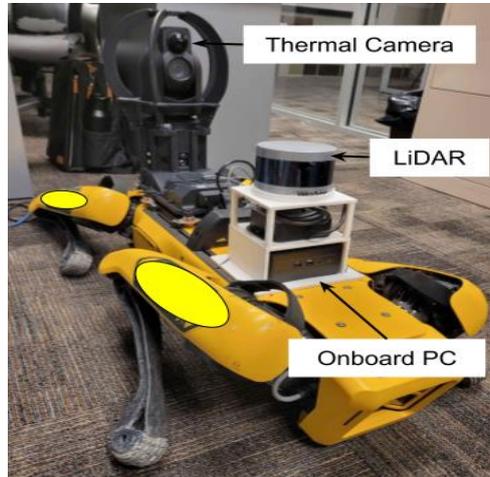

**Fig. 2.** Hardware setup used for the collection of thermal images. Boston Dynamic dog robot with SPOT CAM+IR camera.

### 3.4   Data preprocessing and annotations

Preprocessing is a critical step to ensure the quality and usability of the thermal images. The thermal images captured with Spot CAM+IR were in video stream form, which were then extracted as image frames using a custom Python script. The script utilizes OpenCV framework to extract the frames from the videos. The raw thermal images from the Spot CAM+IR were resized to 640×640 pixels. The images are then annotated using makesense.ai. Fig. 3 shows sample annotated images bounding box ground truth labels used.

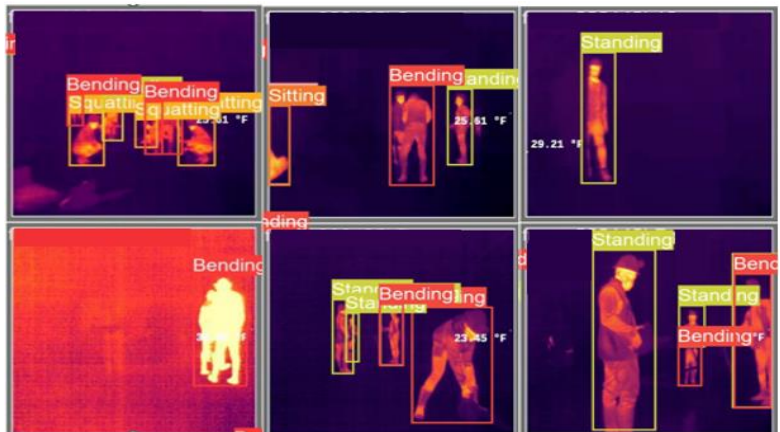

**Fig. 3.** Sample training images with annotated bounding box ground truth labels.

## 4      Object Detection Benchmarks Using Transfer Learning

To validate the dataset's suitability for object detection training, we employed transfer learning with multiple state-of-the-art neural network architectures. We selected YOLOv8 [16], YOLOv10 [17], and YOLO11[18] from the YOLO (You Only Look Once) family for their enhanced feature extraction capabilities and efficient training speed. Additionally, we incorporated RT-DETR (Real-Time Detection Transformer) [19], which leverages transformer architecture to achieve real-time detection while maintaining high accuracy. All experiments utilized data exclusively from the Spot CAM+IR sensor.

The dataset employed for this transfer learning included five (5) posture or object classes, namely, bending, lying, sitting, squatting, and standing. Fig. 4 presents the distribution of object classes of this dataset as identified by their bounding box labels.

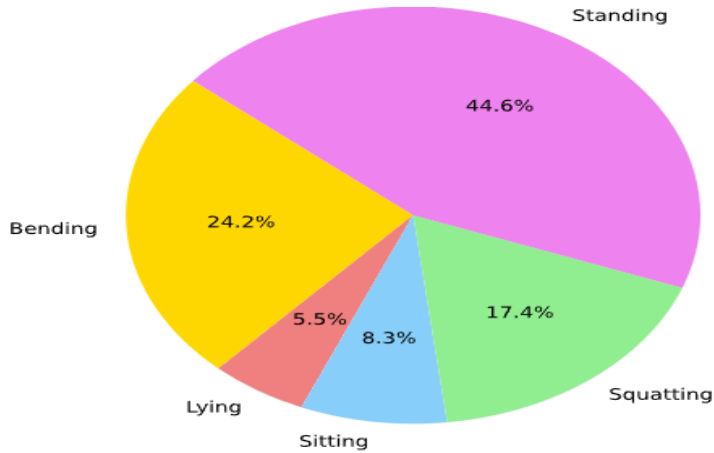

**Fig. 4.** Object class instances distribution in the dataset employed for the study.

**Table 2.** Parameters employed for transfer learning on our thermal dataset for object detection.

| Model / Parameter | Yolov8 | Yolov10 | Yolo11 | RT-DETR |
|---|---|---|---|---|
| Epochs | 300 | 300 | 300 | 300 |
| Learning rate | 0.01 | 0.01 | 0.01 | 0.01 |
| Weight decay | 0.0005 | 0.0005 | 0.0005 | 0.0005 |
| Optimizer | Auto | Auto | SGD | Auto |
| Batch size | 16 | 16 | 16 | 16 |
| Momentum | 0.937 | 0.937 | 0.937 | 0.937 |

For all benchmark experiments presented in the following section, a total of 7,049 thermal images were used in training and validating those models. The dataset was split into training and validation with a ratio of approximately 0.65:0.35, that is 4,584 images for training and 2,465 for validation. Table 2 shows the training parameters such as learning rate, weight decay, optimizer, and momentum. Fig. 5 shows the validation mAP50 learning curves for the best performing models of this study.

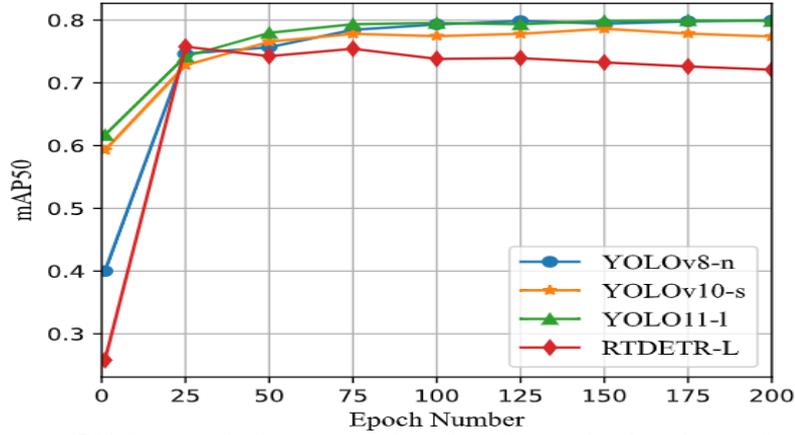

**Fig. 5**. mAP50 dynamics for the entire number of epochs considered for the transfer learning of the best-performing variants. This shows that these models have been sufficiently optimized without overfitting during the transfer learning on our dataset.

## 5    Results and Discussion

We demonstrate the suitability of our dataset on established object detection algorithms like YOLOv8, YOLOv10, YOLO11, and RT-DETR for object detection.

### 5.1    Results

We evaluated the benchmark performance of human detection across previously listed object detection models with and without transfer learning using our proposed dataset. Table 3 shows the comparison of mAP50 and F1 scores.

**Table 3** Results on model performance with and without transfer learning.

| Model | Without transfer learning | | With transfer learning on our dataset | |
|---|---|---|---|---|
| | mAP50 (%) | F1 score (%) | mAP50 (%) | F1 score (%) |
| YOLOv8-x | 61.30 | 64.60 | 78.80 | 85.40 |
| YOLOv10-x | 59.40 | 63.40 | 75.80 | 85.07 |
| YOLO11-x | 57.10 | 61.10 | 79.00 | 85.13 |
| RTDETR-X | 60.70 | 66.00 | 84.80 | 82.61 |

The results after the validation showed that the models performed better after transfer learning with our dataset on YOLOv8-x, YOLOv10x, YOLO11-x, and RTDETR-X. Overall, transfer learning with our dataset improves the models for human detection in an underground mining environment during an emergency.

Although the original dataset has five (5) object classes to represent the various miner postures, for simplicity we initially evaluated the results based on person detection (i.e. all class detection results) as opposed to posture detection. Table 4 shows the mAP, recall, precision, and F1 scores from the study for the detection of all the object classes after the transfer learning.

**Table 4** Results of transfer learning.

| Model | Params | mAP50 (%) | Precision (%) | Recall (%) | F1 score (%) |
|---|---|---|---|---|---|
| YOLOv8-n | 3.0M | 79.90 | 76.20 | 73.20 | 74.70 |
| YOLOv8-s | 11.1M | 79.80 | 77.50 | 71.80 | 74.54 |
| YOLOv8-m | 25.8M | 78.60 | 78.90 | 68.90 | 73.28 |
| YOLOv8-l | 43.6M | 79.00 | 77.50 | 71.60 | 74.43 |
| YOLOv8-x | 68.1M | 78.90 | 76.60 | 71.50 | 73.96 |
| YOLOv10-n | 2.7M | 78.40 | 78.50 | 69.30 | 72.39 |
| YOLOv10-s | 8.0M | 78.80 | 77.80 | 70.90 | 74.19 |
| YOLOv10-m | 16.5M | 78.20 | 75.70 | 71.90 | 73.75 |
| YOLOv10-l | 25.7M | 78.70 | 75.90 | 72.00 | 73.90 |
| YOLOv10-x | 31.6M | 77.80 | 77.30 | 71.60 | 74.34 |
| YOLO11-n | 2.59M | 80.10 | 80.50 | 71.80 | 75.90 |
| YOLO11-s | 9.4M | 80.00 | 78.10 | 72.10 | 74.98 |
| YOLO11-m | 20.1M | 79.40 | 76.80 | 71.90 | 74.27 |
| YOLO11-l | 25.3M | 80.20 | 77.70 | 72.00 | 74.74 |
| YOLO11-x | 56.9M | 79.30 | 77.20 | 71.50 | 74.24 |
| RTDETR-L | 32.0M | 74.80 | 78.50 | 70.30 | 74.17 |
| RTDETR-X | 65.0M | 75.00 | 78.20 | 69.90 | 73.82 |

Clearly, YOLO11-l shows superiority amongst all the other models considered in this study, also two other variants of YOLO11 follow YOLO11-l closes, namely YOLO11-n, and YOLO11-s. Although YOLOv8-n is seen to closely follow the first three variants of YOLO11, it could not outperform them. YOLOv10-s, although the highest among the YOLOv10 variants, is seen to be behind the least mAP50 of YOLOv8. RT-DETRs are seen to be the worst-performing models in terms of mAP50 for this transfer learning study.

From Table 4, the results show that YOLO11-n outperforms all other models in terms of precision, which might not be the trend for YOLO as the largest variant with more trainable parameters is likely to have higher precision [20].

The results show that YOLOv8-n has the highest recall, followed by YOLO11-s, YOLO11-l, and YOLOv10-l, with YOLOv8-m being the least of all. Generally, there is supposed to be an increase in the recall performance as we move from smaller-sized variants of the models to larger-sized variants but none of these models exhibited this trend in the results presented in Table 4.

The results show that YOLO11-n has the highest F1 score, although it did not record the highest mAP50 score. This high F1 score of YOLO11-n could be attributed to the high precision score which harmonizes the effect of a low recall value.

To further show the ability of the dataset to support posture detection, we analyzed the data for the ability to accurately detect the five posture classes. Fig. 6 presents the normalized confusion matrix for this algorithm.

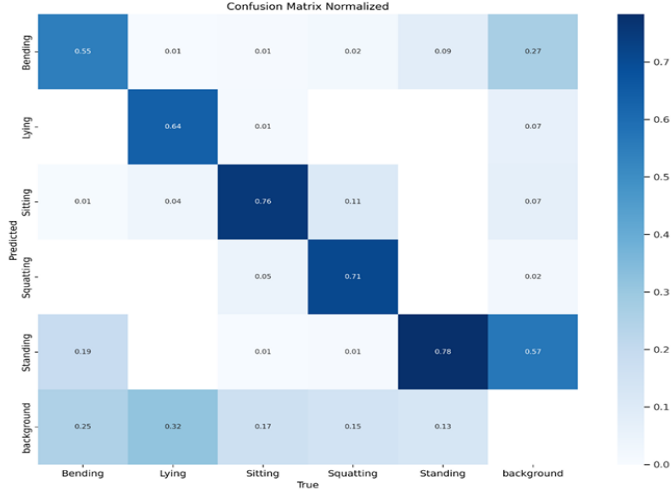

**Fig. 6.** Normalized confusion matrix for one of the YOLO11 variants.

From Fig. 6, the confusion matrix for one of the YOLO11 variants shows bending was wrongly detected as standing, and background falsely detected as standing during our experiment. There were some other false detections for the other posture classes, but they were below 0.15. The detection of bending class was worse as compared to the other classes present in the dataset used for this experiment.

The results still show some misclassifications in detecting certain posture classes after using our dataset for transfer learning. Fig. 7 shows some misclassifications observed after using our dataset for prediction after transfer learning.

### 5.2    Discussion

The results in Table 3 show that transfer learning with our dataset helps improve the model's performance in detecting various posture classes. This could help detect miners in various posture classes during underground mine emergencies to facilitate rescue missions.

The results for the models presented in Table 4 show a diverse array of performance as the accuracy for the object class did not follow any trend. Results for the accuracy, when considering just the ability to detect humans (miners), show that thermal images have different requirements when it comes to object detection. For example, traditionally, the accuracy of the YOLO variants increases as we move from the smallest to the largest variant for most RGB images [20]. However, this is not the case in this experiment. In our experiment, we noted that smaller YOLO model versions regularly demonstrated superior detection performance. This can be ascribed to two primary factors. Firstly, our thermal dataset exhibits a relatively simple class distribution, comprising solely humans in diverse postures. Secondly, the restricted size of our training dataset may advantage smaller structures, a phenomenon previously observed in investigations of object detection utilizing small datasets [21][22].

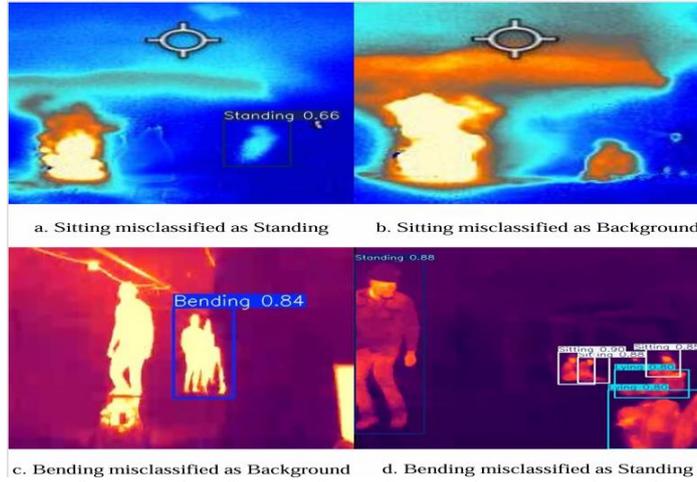

**Fig. 7.** Some misclassification after performing detection with our dataset. Figs 7a. shows sitting misclassified as standing, 7b shows sitting misclassified as background, 7c shows bending misclassified as background, and 7d shows bending misclassified as stand.

While YOLOv11-l attained superior performance in our studies, its enhancement in mAP50 (from 80.1 to 80.2) was negligible when juxtaposed with the lightweight YOLOv11-n variation, despite necessitating tenfold the number of parameters. This indicates that for underground mining applications, where computational resources are significantly limited, the smaller model variant is a more pragmatic alternative, delivering enhanced computational and power efficiency.

The most common mistake that occurs is that the bending class is incorrectly identified as standing. This is most likely the result of an imbalanced training dataset in which standing poses account for 44% of the labels. Future work should concentrate on two most important areas: finding the ideal quantity of extra training data that is required to improve bending detection and improving the detection algorithm itself.

## 6    Conclusion

We presented an image dataset for underground miner detection during emergencies and the results of our evaluation of state-of-the-art algorithms like YOLOv8, YOLOv10, YOLO11, RT-DETR-L, and RT-DETR-X using this dataset. The dataset presented in this work includes 7,049 images acquired with the Boston Dynamics Spot Dog and Spot CAM+IR. The images capture miners in various postures and scenarios probable in an underground mine emergency. To illustrate the usefulness of the data, we conducted transfer learning with YOLOv8, YOLOv10, YOLO11, RT-DETR-L, and RT-DETR-X by splitting the data into 4,584 training and 2,465 validation images. The work shows that YOLO11-l is superior to the other models for underground miner detection. However, we observed misclassifications that could be attributed to class imbalance. This necessitates future work to acquire more images to balance the posture classes.

These would serve as a foundation dataset for miner detection in underground environments during emergencies. Future work will investigate expanding the dataset to include more instances of other posture classes.

**Acknowledgments.** The Centers for Disease Control/ National Occupational Safety and Health (CDC-NIOSH) contract (grant number U60OH012350-01-00) for funding this research.

**Disclosure of Interests.** Authors declare that they do not have any competing interests.

## References

1. Shao, Z., Yan, L., Chen, J., Chen, J.: A Dataset and A Lightweight Object Detection Network for Thermal Image-Based Home Surveillance. In: Proceedings of 2022 Asia-Pacific Signal and Information Processing Association Annual Summit and Conference, APSIPA ASC 2022. pp. 1332–1336. Institute of Electrical and Electronics Engineers Inc. (2022). https://doi.org/10.23919/APSIPAASC55919.2022.9980050.
2. Papageorgiou, C., Poggio, T.: A Trainable System for Object Detection. (2000).
3. Huda, N.U., Hansen, B.D., Gade, R., Moeslund, T.B.: The effect of a diverse dataset for transfer learning in thermal person detection. Sensors (Switzerland). 20, (2020). https://doi.org/10.3390/s20071982.
4. Zhang, Y., Zhou, Y.: YOLOv5 Based Pedestrian Safety Detection in Underground Coal Mines. In: 2021 IEEE International Conference on Robotics and Biomimetics, ROBIO 2021. pp. 1700–1705. Institute of Electrical and Electronics Engineers Inc. (2021). https://doi.org/10.1109/ROBIO54168.2021.9739594.
5. Dang, T., Mascarich, F., Khattak, S., Nguyen, H., Nguyen, H., Hirsh, S., Papachristos, C., Alexis, K.: Autonomous Search for Underground Mine Rescue Using Aerial Robots. In: 2020 IEEE Aerospace Conference. IEEE (2020).
6. Dickens, J.S., van Wyk, M.A., Green, J.J.: Pedestrian detection for underground mine vehicles using thermal images. In: IEEE Africon'11. pp. 1–6. Institute of Electrical and Electronics Engineers (2011).
7. Szrek, J., Zimroz, R., Wodecki, J., Michalak, A., Góralczyk, M., Worsa-Kozak, M.: Application of the infrared thermography and unmanned ground vehicle for rescue action support in underground mine—the amicos project. Remote Sens (Basel). 13, 1–20 (2021). https://doi.org/10.3390/rs13010069.
8. Maningo, J.M., Amoroso, M.C.C., Atienza, K.R., Ladera, R.K., Menodiado, N.M., Ambata, L.U., Cabatuan, M.K., Sybingco, E., Bandala, A., Española, J., Vicerra, R.R.: Thermal Imaging Dataset for Human Presence Detection. In: 2023 8th International Conference on Business and Industrial Research, ICBIR 2023 - Proceedings. pp. 1159–1164. Institute of Electrical and Electronics Engineers Inc. (2023). https://doi.org/10.1109/ICBIR57571.2023.10147404.
9. Zhu, S., Voigt, T., Perez-Ramirez, D.F., Eriksson, J.: Dataset: A Low-resolution infrared thermal dataset and potential privacy-preserving applications. In: SenSys 2021 - Proceedings of the 2021 19th ACM Conference on Embedded Networked Sensor Systems. pp. 552–555. Association for Computing Machinery, Inc (2021). https://doi.org/10.1145/3485730.3493692.
10. Berg, Amanda.: Detection and Tracking in Thermal Infrared Imagery. Linkopings Universitet (2016).
11. Rizk, M., Bayad, I.: Human Detection in Thermal Images Using YOLOv8 for Search and Rescue Missions. In: International Conference on Advances in Biomedical Engineering, ICABME. pp. 210–215. Institute of Electrical and Electronics Engineers Inc. (2023). https://doi.org/10.1109/ICABME59496.2023.10293139.
12. Bondi, E., Jain, R., Aggrawal, P., Anand, S., Hannaford, R., Kapoor, A., Piavis, J., Shah, S., Joppa, L., Dilkina, B., Tambe, M.: BIRDSAI: A Dataset for Detection and Tracking in Aerial Thermal Infrared Videos. In: Proceedings of the IEEE/CVF winter conference on applications of computer vision. pp. 1747–1756 (2020).
13. Cerutti, G., Milosevic, B., Farella, E.: Outdoor People Detection in Low Resolution Thermal Images. In: 2018 3rd International Conference on Smart and Sustainable Technologies (SpliTech). pp. 1–6. IEEE (2018).
14. Liu, Q., He, Z., Li, X., Zheng, Y.: PTB-TIR: A Thermal Infrared Pedestrian Tracking Benchmark. IEEE Trans Multimedia. 22, 666–675 (2020). https://doi.org/10.1109/TMM.2019.2932615.
15. Krišto, M., Ivašić-Kos, M.: Thermal Imaging Dataset for Person Detection. In: 2019 42nd International Convention on Information and Communication Technology, Electronics and Microelectronics (MIPRO). pp. 1126–1131. IEEE (2019).
16. Jocher, G., Chaurasia, A., Qiu, J.: Ultralytics YOLOv8, https://github.com/ultralytics/ultralytics, (2023).
17. Wang, A., Chen, H., Liu, L., Chen, K., Lin, Z., Han, J., Ding, G.: YOLOv10: Real-Time End-to-End Object Detection. arXiv preprint arXiv:2405.14458. 1–21 (2024).
18. Jocher, G., Qiu, J.: Ultralytics YOLO11, https://github.com/ultralytics/ultralytics, (2024).
19. Zhao, Y., Lv, W., Xu, S., Wei, J., Wang, G., Dang, Q., Liu, Y., Chen, J.: DETRs Beat YOLOs on Real-time Object Detection. (2023).
20. Hussain, M.: YOLOv5, YOLOv8 and YOLOv10: The Go-To Detectors for Real-time Vision. (2024).
21. Gamani, A.-R.A., Arhin, I., Asamoah, A.K.: Performance Evaluation of YOLOv8 Model Configurations, for Instance Segmentation of Strawberry Fruit Development Stages in an Open Field Environment. (2024).

22. Xiao, B., Nguyen, M., Yan, W.Q.: Fruit ripeness identification using YOLOv8 model. Multimed Tools Appl. 83, 28039–28056 (2024). https://doi.org/10.1007/s11042-023-16570-9.